\documentclass[letterpaper]{article} % DO NOT CHANGE THIS
\usepackage[preprint]{aaai2027}  % DO NOT CHANGE THIS
\usepackage[hyphens]{url}  % DO NOT CHANGE THIS
\usepackage{graphicx} % DO NOT CHANGE THIS
\usepackage{natbib}  % DO NOT CHANGE THIS AND DO NOT ADD ANY OPTIONS TO IT
\usepackage{caption} % DO NOT CHANGE THIS AND DO NOT ADD ANY OPTIONS TO IT
\usepackage{algorithm}
\usepackage{algorithmic}

\usepackage{newfloat}
\usepackage{listings}
\DeclareCaptionStyle{ruled}{labelfont=normalfont,labelsep=colon,strut=off} % DO NOT CHANGE THIS
\floatstyle{ruled}
\newfloat{listing}{tb}{lst}{}
\floatname{listing}{Listing}

\usepackage{booktabs}

\usepackage{amsmath}
\usepackage{booktabs}
\usepackage{multirow}
\usepackage{xcolor}
\usepackage{colortbl}

\title{TopoAgent: A Self-Evolving Topological Agent for Multimodal Scientific Reasoning}
\author{
    Mingze Xu\textsuperscript{\rm 1},
    Yinghui Li\textsuperscript{\rm 1},
    Jiayi Kuang\textsuperscript{\rm 2}, Zhanhui Kang\textsuperscript{\rm 3}, Di Yin\textsuperscript{\rm 3} \\ Ying Shen\textsuperscript{\rm 2}, Xing Sun\textsuperscript{\rm 3}, Yuxing Han\textsuperscript{\rm 1}\\
}
\affiliations{
    \textsuperscript{\rm 1}Tsinghua University\\
    \textsuperscript{\rm 2}Sun Yat-sen University, 
    \textsuperscript{\rm 3}Tencent\\
    liyinghuihhh@gmail.com
}

\begin{document}

\maketitle

\begin{abstract}
While Multimodal Large Language Models (MLLMs) excel in general tasks, rigorous scientific reasoning remains challenging due to the limitations of monolithic, linear planning. Such sequential designs often suffer from visual-semantic misalignment, long-context hallucinations, and brittle execution under fixed task granularity. We propose \textbf{TopoAgent}, a self-evolving topological framework that replaces linear trajectories with dynamic, state-isolated graph evolution. TopoAgent first employs a front-end decomposer to fracture complex queries into visually-grounded atoms. These atoms are organized into a Directed Acyclic Graph (DAG) based on their dependencies, enabling strict context isolation to shield the reasoning engine from irrelevant historical noise. Furthermore, we introduce adaptive atomic fission, which dynamically splits bottleneck nodes into finer-grained sub-atoms at runtime when tool capability boundaries are exceeded. Extensive experiments across mathematics, physics, and chemistry benchmarks demonstrate that TopoAgent significantly outperforms state-of-the-art linear agent frameworks, providing a robust, noise-resistant, and self-correcting paradigm for autonomous scientific reasoning.
\end{abstract}


\section{Introduction}
\label{sec:intro}

The advent of Multimodal Large Language Models (MLLMs) has catalyzed significant progress \cite{kuang2025natural, an2026toward} in developing autonomous agents capable of interacting with complex, unstructured environments \cite{lu2025youtu, kuang2025process, du2026survey, zhang2026chatbot, liu2026tangrampuzzle}. Recently, multimodal scientific reasoning---spanning Mathematics, Physics, and Chemistry---has emerged as the ultimate crucible for these agents \cite{wang2026frontierscience, liu2025atlas, lupidi2026airs}. Unlike general visual question answering, scientific reasoning demands a rigorous synergy between fine-grained visual perception and deep symbolic deduction \cite{li2026cognitive, burgess2025microvqa}. To navigate this complexity, existing agentic frameworks predominantly rely on linear planning paradigms, such as Chain-of-Thought (CoT) \cite{wei2022chain} or sequential Plan-and-Solve strategies \cite{yao2022react}, wherein the model generates a macroscopic step-by-step trajectory and accumulates the entire execution history as context for subsequent actions.

However, deploying linear agentic frameworks in dense scientific domains exposes three fundamental architectural flaws. First, \textbf{Cognitive Overload and Misalignment}: Existing agents typically intertwine raw visual perception with abstract logical calculation within a single generation step. This premature convergence often leads to critical visual variables being hallucinated or ignored \cite{li2026cognitive, luo2026p1}. Second, \textbf{Long-Context Hallucination via Linear Accumulation}: Scientific problem-solving is inherently non-linear. Extracting independent variables constitutes parallel sub-processes. Forcing these topologically independent steps into a strict linear sequence, while blindly concatenating all historical outputs into the prompt, injects irrelevant noise and fundamentally corrupts the model's attention mechanism during downstream equation solving. Third, \textbf{Static Capability Horizons}: Current planners generate static execution steps with fixed granularity. If a scheduled task exceeds the capability boundary of the assigned tool, the entire pipeline typically deadlocks or aborts, lacking the resilience to dynamically adapt its reasoning resolution.

To surmount these fundamental bottlenecks, we propose \textbf{TopoAgent}, a self-evolving topological framework driven by a Directed Acyclic Graph (DAG) of dynamic atomic executions. TopoAgent fundamentally shifts the autonomous reasoning paradigm from linear trajectory accumulation to non-linear, state-isolated topological evolution. Our framework initially employs a front-end \textit{Decomposer} to fracture complex queries into visually-grounded, perception-first \textit{Atoms}--- minimal executable units that bind concepts to explicit visual regions before logical computation begins.

\begin{figure}[t]
    \centering
    \includegraphics[width=0.65\linewidth]{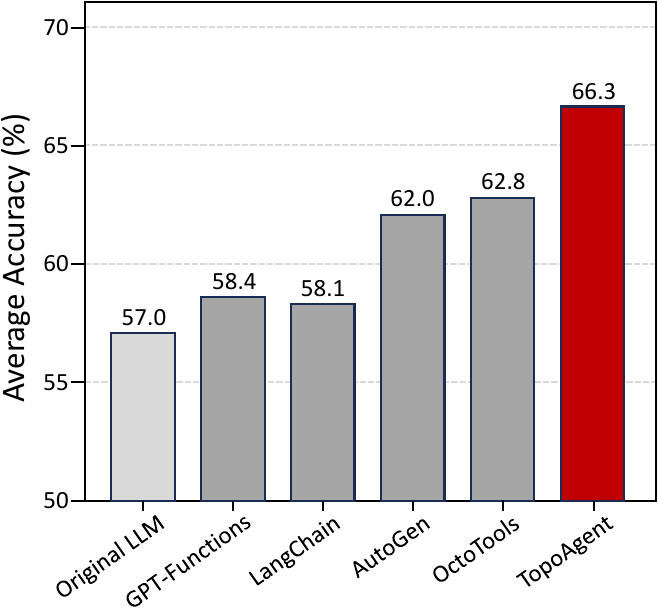}
    \caption{Comparison of global average accuracy across six evaluated MLLMs on multimodal scientific reasoning benchmarks. TopoAgent achieves a state-of-the-art overall performance of 66.3\%, outperforming standard LLMs, sequential planners, and contemporary agentic frameworks.}
    \label{fig:zongtifenshu}
\end{figure}

Crucially, instead of executing these atoms as a flat list, TopoAgent orchestrates them into a Directed Acyclic Graph based on their intrinsic prerequisite relationships. This topology enables \textit{Context Isolation}: when executing a logical node, the agent only ingests the highly-purified, deterministic results from its direct prerequisite nodes, completely shielding the reasoning engine from irrelevant historical noise. Furthermore, to address the static granularity problem, we introduce \textit{Adaptive Atomic Fission}. During runtime, if an atom cannot be deterministically resolved within a strict horizon, the framework dynamically fractures this bottleneck node into finer-grained sub-atoms and prepends them to a dynamic execution queue. This hierarchical namespace mapping allows the agent to iteratively lower the task difficulty to match its tool capabilities without recompiling the global graph.

We conduct extensive experiments across a suite of rigorous multimodal scientific benchmarks encompassing Mathematics, Physics, and Chemistry. Comprehensive evaluations demonstrate that TopoAgent significantly outperforms existing state-of-the-art agent frameworks. As illustrated in Figure \ref{fig:zongtifenshu}, TopoAgent achieves a leading global average accuracy of 66.3\% across six diverse foundation models, establishing a substantial margin over both linear planners and multi-agent systems. By formalizing perception-reasoning decoupling and dynamic graph evolution, TopoAgent establishes a robust pathway toward reliable scientific discovery.

In summary, our main contributions are three-fold:
\begin{itemize}
    \item We identify the structural limitations of linear agentic planning in multimodal scientific reasoning and propose \textbf{TopoAgent}, a self-evolving topological framework that transforms sequential planning into a Directed Acyclic Graph (DAG) of visually-grounded atomic tasks.
    \item We introduce a strict \textit{Context Isolation} mechanism governed by DAG topology, which effectively eliminates long-context hallucinations by routing only highly-purified prerequisite states to downstream logical nodes.
    \item We design an \textit{Adaptive Atomic Fission} mechanism that endows the agent with self-evolutionary resilience, allowing it to dynamically degrade task granularity at runtime to match underlying tool capabilities.
\end{itemize}

\section{Related Work}

\subsection{Multimodal Scientific Reasoning}
Recent MLLM benchmarks emphasize rigorous reasoning in mathematics \cite{lu2023mathvista, peng2024multimath, li2025one, kuang2026atomic}, physics \cite{yue2024mmmu, he2024olympiadbench}, and chemistry \cite{li2026cognitive}, demanding deep synthesis of multimodal elements \cite{li2024towards, zhou2025scientists, li2022past, goswami2025chartcitor, kuang2025express,li2025benchmarking}. However, MLLMs often suffer from visual-semantic misalignment, attempting logical deduction before securely anchoring visual variables. For Olympiad-level problems that overwhelm inherent parametric knowledge \cite{cui2025evaluating, sun2025challenging, yu2025hipho}, tool-augmented and structured reasoning paradigms have become essential.

\subsection{Tool-Augmented and Topological Agents}
To extend model capabilities, autonomous agents offload specialized subtasks to external tools \cite{ye2025productagent, li2025towards, schick2023toolformer, qin2023toolllm, li2025sciagent, yu2025physicsminions, huang2026deep, guo2026evoconfig}. While sequential frameworks \cite{wu2024autogen, bran2023chemcrow} based on CoT \cite{wei2022chain} or ReAct \cite{yao2022react} are prevalent, they accumulate execution history that injects noise and triggers error propagation. Advanced topologies like Tree-of-Thoughts \cite{yao2023tree} and Graph-of-Thoughts \cite{besta2024graph} address long-horizon planning but remain inadequate for scientific tasks due to two core vulnerabilities: (1) lack of \textbf{context isolation}, where redundant data from parallel branches corrupts the deterministic states required for deduction; and (2) \textbf{static task granularity}, which leads to deadlocks when a sub-task exceeds tool capabilities. Our TopoAgent surmounts these by enabling state-isolated evolution and dynamic task fracturing.

\begin{figure*}[!t]
    \centering
    \includegraphics[width=0.9\linewidth]{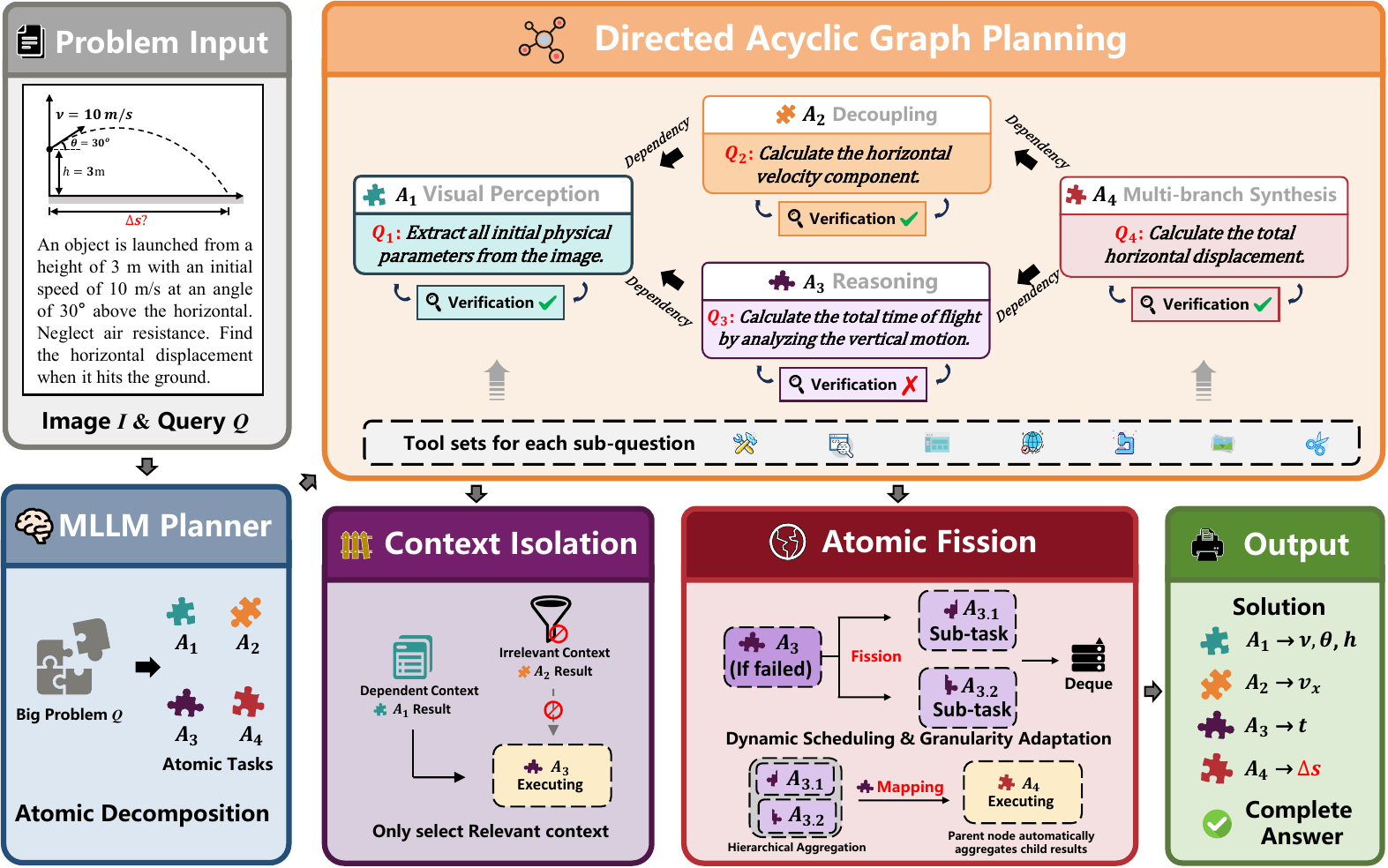}
    \caption{The overall architecture of the proposed TopoAgent framework. Given a complex multimodal scientific query, the framework first performs \textbf{Visually-Grounded Atomic Decomposition} to break the problem into fine-grained, executable atoms. These atoms are then orchestrated via \textbf{Directed Acyclic Graph (DAG) Planning}, which enforces strict \textbf{Context Isolation} by routing only relevant prerequisite states to downstream nodes, thereby mitigating hallucination. During execution, if an atom exceeds the tool's capability boundary, the \textbf{Adaptive Atomic Fission} mechanism dynamically splits it into finer-grained sub-tasks at runtime, enabling self-evolving and robust reasoning.}
    \label{fig:overview}
\end{figure*}

\section{Methodology}
\label{sec:method}

To address the limitations of monolithic planning in MLLMs, we propose \textbf{TopoAgent}. Existing agents often entangle visual perception with logical deduction within a single window, leading to error accumulation and premature logical convergence. TopoAgent decouples these processes by shifting from sequential prompting to graph-based atomic execution. As illustrated in Figure \ref{fig:overview}, the architecture comprises three core mechanisms: (1) \textbf{Visually-Grounded Atomic Decomposition}, which translates queries into perception-first units; (2) \textbf{Directed Acyclic Graph (DAG) Planning} for strict context isolation; and (3) \textbf{Adaptive Atomic Fission} for dynamic granularity adjustment at runtime.

\subsection{Visually-Grounded Atomic Decomposition}
\label{subsec:atomic_decomp}

Traditional agents rely on holistic policies $\pi(a_t \mid Q, I, h_{<t})$ that often suffer from visual-semantic misalignment as they attempt to synthesize theorems before anchoring visual variables. To overcome this, our \textit{Decomposer} module $\mathcal{D}$ fractures the problem into discrete, visually-grounded \textit{Atoms} $\mathcal{A} = \{A_1, A_2, \ldots, A_n\}$. 

Each atom $A_i$ is a minimal solvable unit encapsulating a local objective and an expected symbolic state. Crucially, this decomposition is \textit{perception-first}: the Decomposer prioritizes the spatial grounding of visual elements before any logical deduction begins. By mandating this step, TopoAgent forces the MLLM to anchor abstract concepts to deterministic symbolic representations, significantly mitigating multimodal hallucinations in subsequent reasoning phases.

\begin{algorithm}[!t]
\caption{Dynamic Execution of the TopoAgent Framework}
\label{alg:dag_atom}
\begin{algorithmic}[1]

\STATE \textbf{Input:} Complex query $Q$, image $I$, toolset $T$, maximum execution steps $T_{\max}$
\STATE \textbf{Output:} Final answer $Ans$

\STATE Initialize memory module $\mathcal{M}\leftarrow\emptyset$
\STATE Initialize execution queue $\mathcal{Q}_{exec}\leftarrow\emptyset$

\STATE $\mathcal{G}(\mathcal{A},\mathcal{E})\leftarrow Decomposition(Q,I)$
\STATE Push $\text{TopologicalSort}(\mathcal{G})$ into $\mathcal{Q}_{exec}$

\WHILE{$\mathcal{Q}_{exec}\neq\emptyset$}

    \STATE $A_i\leftarrow \mathcal{Q}_{exec}.\mathrm{pop\_front}()$
    \STATE $\mathcal{C}(A_i)\leftarrow
    \displaystyle\bigcup_{A_j\in\mathcal{P}(A_i)}\mathcal{M}(A_j)$

    \STATE $\mathrm{is\_resolved}\leftarrow \mathrm{False}$
    \STATE $\mathrm{trace}\leftarrow\emptyset$

    \FOR{$t=1$ \TO $T_{\max}$}

        \STATE $(act,t_{select})\leftarrow
        Planner(A_i,\mathcal{C}(A_i),I)$

        \STATE $res\leftarrow Execute(act,t_{select}\in T)$

        \STATE Update $\mathcal{M}(A_i)$ with $res$

        \STATE Append $(act,res)$ to trace

        \IF{$Verifier(A_i,res)=STOP$}
            \STATE $\mathrm{is\_resolved}\leftarrow \mathrm{True}$
            \STATE \textbf{break}
        \ENDIF

    \ENDFOR

    \IF{not $\mathrm{is\_resolved}$}

        \STATE $\{A_{i.1},\ldots,A_{i.k}\}\leftarrow
        \mathcal{F}(A_i,\mathrm{trace})$

        \STATE Push $\{A_{i.1},\ldots,A_{i.k}\}$ to the front of $\mathcal{Q}_{exec}$

        \STATE Update $\mathcal{G}(\mathcal{A},\mathcal{E})$
        with hierarchical namespace mappings

    \ENDIF

\ENDWHILE

\STATE $Ans\leftarrow Solver(Q,\mathcal{M})$
\STATE \textbf{return} $Ans$

\end{algorithmic}
\end{algorithm}

\subsection{Directed Acyclic Graph Planning and Context Isolation}
\label{subsec:dag_planning}

Complex multimodal reasoning tasks are inherently non-linear. Extracting independent visual features often involves parallel sub-processes, while downstream deductions require their confluence. Conventional linear frameworks sequentially accumulate trajectories, forcing MLLMs to process dense, irrelevant contexts, which exacerbates attention dilution and long-context hallucinations.

To mitigate this, we formalize atomic planning as a directed acyclic graph (DAG) $\mathcal{G} = (\mathcal{A}, \mathcal{E})$. Nodes $\mathcal{A} = \{A_1, \dots, A_n\}$ represent atomic sub-tasks generated by the Decomposer, and edges $\mathcal{E}$ define prerequisite dependencies. The prerequisite set for atom $A_i$ is defined as $\mathcal{P}(A_i) = \{A_j \mid (A_j, A_i) \in \mathcal{E}\}$.

During execution, we introduce a \textit{Context Isolation} mechanism. Let $\mathcal{M}$ be the dynamic memory storing the states of completed atoms. When scheduling $A_i$, instead of injecting the entire history, the system constructs a purified local context $\mathcal{C}(A_i)$:
\begin{equation}
    \mathcal{C}(A_i) = \bigcup_{A_j \in \mathcal{P}(A_i)} \mathcal{M}(A_j)
\end{equation}
This topological isolation shields the reasoning engine from irrelevant historical noise, ensuring variables are precisely bound to their corresponding symbols and preventing multimodal feature cross-contamination.

\subsection{Adaptive Atomic Fission for Dynamic Evolution}
\label{subsec:atomic_fission}

A key bottleneck in static DAG planning is the mismatch between task granularity and tool capabilities. If an atom $A_i$ is too complex for single-step execution, standard frameworks exhaust retry limits and fail. To provide resilience, we design the \textit{Adaptive Atomic Fission} mechanism.

We manage execution via a dynamic double-ended queue (deque) $\mathcal{Q}_{exec}$. If the agent fails to reach a ``STOP'' state for an active atom $A_i$ within $T_{max}$ steps, $A_i$ is flagged as a \textit{capability-exceeding node}.

Rather than aborting the global process, the framework triggers a fission function $\mathcal{F}$. Using the LLM's reflexive capability, $\mathcal{F}$ analyzes the failed trace and fractures $A_i$ into finer-grained sub-atoms:
\begin{equation}
    \mathcal{F}(A_i, \text{trace}) \rightarrow \{A_{i.1}, A_{i.2}, \dots, A_{i.k}\}
\end{equation}
These sub-atoms are topologically sorted and prepended to $\mathcal{Q}_{exec}$ to prioritize resolving this local bottleneck.

To preserve global DAG integrity without recompilation, we introduce \textit{Hierarchical Dependency Mapping}. Descendants $A_{i.*}$ inherit the namespace of their parent $A_i$. When a subsequent atom queries $\mathcal{M}$ for $A_i$, the retrieval function automatically aggregates results across the sub-atomic namespace:
\begin{equation}
    \mathcal{M}_{virtual}(A_i) = \bigoplus_{k} \mathcal{M}(A_{i.k})
\end{equation}

This hierarchical aggregation decouples local granularity adjustments from the global topology, enabling dynamic capability-matching and robust problem-solving even with imperfect initial decompositions.

Algorithm~\ref{alg:dag_atom} summarizes the complete TopoAgent execution, highlighting the interplay between context isolation and dynamic atomic fission.

\section{Experiments}

\begin{table*}[t]
\centering
\caption{Comprehensive performance comparison of TopoAgent against baseline frameworks. To provide a rigorous evaluation, we detail the results using the most advanced proprietary model (\textbf{GPT-5}) and open-weight model (\textbf{Qwen3-vl}), followed by the \textbf{Global Average} across all six evaluated MLLMs. \textit{Detailed performance metrics for the remaining four MLLMs are provided in the Appendix.} The best result in each column per block is highlighted in \textbf{bold}.}
\label{tab:main_results}
% 定义一个命令用于快速生成灰色小号的标准差
\newcommand{\sd}[1]{\textcolor{gray}{\scriptsize{$\pm$#1}}}

\resizebox{0.9\textwidth}{!}{%
\begin{tabular}{@{} l ccc cccc c @{}}
\toprule
\multirow{2}{*}{\textbf{Framework}} & \multicolumn{3}{c}{\textbf{Mathematics}} & \multicolumn{3}{c}{\textbf{Physics}} & \multicolumn{1}{c}{\textbf{Chemistry}} & \multirow{2}{*}{\textbf{AVG.}} \\
\cmidrule(lr){2-4} \cmidrule(lr){5-7} \cmidrule(lr){8-8}
& \textbf{MathVista} & \textbf{Multi-math} & \textbf{Math-Sym} & \textbf{Olympiad} & \textbf{MMMU} & \textbf{Phy-Sym} & \textbf{Chem-Sym} & \\
\midrule

% ==================== BLOCK 1: GPT-5 ====================
\rowcolor{gray!15} \multicolumn{9}{l}{\textbf{\textsc{Backbone: GPT-5}} (Representative Proprietary MLLM)} \\
\midrule
Original LLM  & 77.8 \sd{0.3} & 64.0 \sd{2.1} & 68.3 \sd{0.8} & 45.8 \sd{2.5} & 65.0 \sd{1.9} & 57.7 \sd{2.3} & 53.5 \sd{2.0} & 61.7 \\
OctoTools     & 72.2 \sd{1.5} & \textbf{78.5} \sd{0.8} & 73.5 \sd{2.1} & 55.3 \sd{2.8} & \textbf{85.8} \sd{0.1} & 70.8 \sd{1.7} & 59.7 \sd{2.4} & 70.8 \\
LangChain     & 78.2 \sd{0.3} & 73.7 \sd{1.9} & 74.0 \sd{0.5} & \textbf{64.7} \sd{3.1} & 81.2 \sd{1.6} & 71.8 \sd{2.0} & 51.8 \sd{2.7} & 70.8 \\
GPT-Functions & 71.2 \sd{2.2} & 75.8 \sd{1.4} & 71.0 \sd{1.8} & 40.2 \sd{3.0} & 71.3 \sd{0.3} & 53.5 \sd{2.6} & 56.0 \sd{1.9} & 62.7 \\
AutoGen       & \textbf{80.2} \sd{1.1} & 72.0 \sd{2.0} & 73.3 \sd{1.7} & 55.8 \sd{2.5} & 84.0 \sd{0.3} & 67.3 \sd{2.1} & 54.5 \sd{2.3} & 69.6 \\
\textbf{TopoAgent (Ours)} & 78.3 \sd{1.4} & 74.3 \sd{1.6} & \textbf{74.5} \sd{0.2} & 60.3 \sd{2.2} & 84.5 \sd{1.3} & \textbf{72.7} \sd{1.5} & \textbf{60.0} \sd{1.8} & \textbf{72.1} \\
\midrule

% ==================== BLOCK 2: Qwen3 ====================
\rowcolor{gray!15} \multicolumn{9}{l}{\textbf{\textsc{Backbone: Qwen3-vl-235B-A22B-Instruct}} (Representative Open-Weight MLLM)} \\
\midrule
Original LLM  & 68.2 \sd{2.1} & 79.7 \sd{0.5} & 72.5 \sd{1.9} & 24.7 \sd{2.8} & 73.5 \sd{0.7} & 49.0 \sd{2.4} & 47.8 \sd{2.2} & 59.3 \\
OctoTools     & 73.3 \sd{1.8} & 76.0 \sd{0.0} & 70.5 \sd{2.2} & 36.2 \sd{3.1} & 77.8 \sd{1.6} & 55.5 \sd{2.1} & 59.8 \sd{1.8} & 64.2 \\
LangChain     & \textbf{85.8} \sd{1.2} & \textbf{82.2} \sd{1.4} & \textbf{80.8} \sd{0.3} & 33.8 \sd{2.5} & 74.2 \sd{1.9} & 53.7 \sd{2.6} & 36.3 \sd{2.9} & 63.8 \\
GPT-Functions & 80.2 \sd{1.6} & 81.8 \sd{0.5} & 77.3 \sd{1.8} & 32.3 \sd{2.7} & \textbf{81.3} \sd{1.4} & 56.3 \sd{2.0} & 38.0 \sd{3.2} & 63.9 \\
AutoGen       & 82.0 \sd{0.3} & 75.8 \sd{2.1} & 76.7 \sd{1.7} & 27.2 \sd{2.9} & 74.3 \sd{2.2} & 50.0 \sd{2.5} & 49.8 \sd{2.3} & 62.3 \\
\textbf{TopoAgent (Ours)} & 80.5 \sd{1.5} & 80.3 \sd{1.7} & 78.7 \sd{1.4} & \textbf{42.7} \sd{2.4} & 75.8 \sd{1.8} & \textbf{56.3} \sd{1.9} & \textbf{63.7} \sd{0.6} & \textbf{68.3} \\
\midrule

% ==================== BLOCK 3: ALL AVERAGE ====================
\rowcolor{gray!15} \multicolumn{9}{l}{\textbf{\textsc{Global Average}} (Aggregated across all 6 evaluated MLLMs)} \\
\midrule
Original LLM  & 73.7 & 71.8 & 70.3 & 30.9 & 62.4 & 45.5 & 44.4 & 57.0 \\
OctoTools     & 74.2 & 73.7 & 71.4 & 35.7 & 72.4 & 52.0 & 60.5 & 62.8 \\
LangChain     & 75.8 & 69.4 & 70.4 & 34.0 & 63.9 & 47.6 & 45.9 & 58.1 \\
GPT-Functions & 78.1 & 74.9 & 73.9 & 26.2 & 66.3 & 44.4 & 44.8 & 58.4 \\
AutoGen       & 77.1 & 75.0 & 73.8 & 35.6 & 69.8 & 51.0 & 51.9 & 62.0 \\
\textbf{TopoAgent (Ours)} & \textbf{79.2} & \textbf{76.3} & \textbf{75.9} & \textbf{41.2} & \textbf{73.6} & \textbf{55.1} & \textbf{62.6} & \textbf{66.3} \\
\bottomrule
\end{tabular}%
}
\end{table*}

\subsection{Experimental Setup}

\textbf{Datasets.} We select challenging benchmarks across three core scientific domains:
\begin{itemize}
    \item \textbf{Mathematics:} \textit{MathVista} ~\cite{lu2023mathvista}, \textit{Multi-math} ~\cite{peng2024multimath}, and the mathematics subset of SymbolBench ~\cite{li2026cognitive} (denoted as Math-Symbol).
    \item \textbf{Physics:} \textit{OlympiadBench}~\cite{he2024olympiadbench}, the physics-related subset of \textit{MMMU}~\cite{yue2024mmmu} (encompassing Physics, Mechanical, Energy \& Power, and the physics subset of SymbolBench~\cite{li2026cognitive} (denoted as Physics-Symbol).
    \item \textbf{Chemistry:}  The chemistry subset of SymbolBench~\cite{li2026cognitive} (denoted as Chem-Symbol).
\end{itemize}

\textbf{Models.} To ensure the generalizability of our evaluation, we instantiate our framework and all baselines using a diverse set of state-of-the-art Multimodal Large Language Models (MLLMs), including \texttt{GPT-5}\cite{5}, \texttt{GPT-5-mini}\cite{5-mini}, \texttt{Gemini-3.1-\allowbreak Pro-Preview}\cite{gemini}, \texttt{Claude-\allowbreak Sonnet-\allowbreak 4-5}\cite{claude}, \texttt{Qwen3-\allowbreak vl-\allowbreak 235B-A22B-Instruct} \cite{bai2025qwen3}, and \texttt{GLM-\allowbreak 4.5V-\allowbreak 106b-\allowbreak a12b} \cite{hong2025glm}.

\textbf{Baselines.} We compare our proposed TopoAgent against several dominant reasoning paradigms and agent frameworks: 
(1) \textbf{Original LLM}: Direct reasoning and answering without external tool orchestration or agentic planning; 
(2) \textbf{OctoTools} \cite{lu2025octotools}: A contemporary tool-augmented agent framework; 
(3) \textbf{LangChain}\cite{langchain2024}: A popular sequential planning framework (using its graph-based execution variant); 
(4) \textbf{GPT-Functions}\cite{gptf}: The standard OpenAI function-calling pipeline; 
(5) \textbf{AutoGen} \cite{wu2024autogen}: A multi-agent conversational framework.

\begin{table*}[!t]
\centering
\caption{Ablation study of TopoAgent based on the Global Average performance across 6 MLLMs. ``w/o DAG Planning'' replaces topological context isolation with linear accumulation. ``w/o Atomic Fission'' disables dynamic task granularity degradation.}
\label{tab:ablation}
\resizebox{0.9\textwidth}{!}{
\begin{tabular}{lcccccccc}
\toprule
\multirow{2}{*}{\textbf{Model Variant}} & \multicolumn{3}{c}{\textbf{Mathematics}} & \multicolumn{3}{c}{\textbf{Physics}} & \textbf{Chemistry} & \multirow{2}{*}{\textbf{AVG.}} \\
\cmidrule(lr){2-4} \cmidrule(lr){5-7} \cmidrule(lr){8-8}
& \textbf{MathVista} & \textbf{Multi-math} & \textbf{Math-Sym} & \textbf{Olympiad} & \textbf{MMMU} & \textbf{Phy-Sym} & \textbf{Chem-Sym} & \\
\midrule
\textbf{Full TopoAgent (Ours)} & \textbf{79.2} & \textbf{76.3} & \textbf{75.9} & \textbf{41.2} & \textbf{73.6} & \textbf{55.1} & \textbf{62.6} & \textbf{66.3} \\
\midrule
w/o DAG Planning & 77.1 & 75.0 & 74.1 & 39.6 & 72.4 & 53.8 & 62.1 & 64.9 \\
\textit{Performance Drop} & \textit{\textcolor{red}{-2.1}} & \textit{\textcolor{red}{-1.3}} & \textit{\textcolor{red}{-1.8}} & \textit{\textcolor{red}{-1.6}} & \textit{\textcolor{red}{-1.2}} & \textit{\textcolor{red}{-1.3}} & \textit{\textcolor{red}{-0.5}} & \textit{\textcolor{red}{-1.4}} \\
\midrule
w/o Atomic Fission & 78.3 & 75.6 & 75.1 & 40.6 & 73.1 & 54.5 & 62.5 & 65.7 \\
\textit{Performance Drop} & \textit{\textcolor{red}{-0.9}} & \textit{\textcolor{red}{-0.7}} & \textit{\textcolor{red}{-0.8}} & \textit{\textcolor{red}{-0.6}} & \textit{\textcolor{red}{-0.5}} & \textit{\textcolor{red}{-0.6}} & \textit{\textcolor{red}{-0.1}} & \textit{\textcolor{red}{-0.6}} \\
\bottomrule
\end{tabular}
}
\end{table*}

\subsection{Main Results}
Table~\ref{tab:main_results} details performance of GPT-5, Qwen3-vl, and global average across six MLLMs; full metrics are in the Appendix.

\textbf{Overall Superiority.} TopoAgent achieves a SOTA global average accuracy of 66.3\%, significantly outperforming sequential planners (LangChain, 58.1\%) and multi-agent frameworks (AutoGen, 62.0\%). This margin confirms that decoupling perception from reasoning via DAG planning effectively mitigates hallucinations and visual-semantic misalignments common in linear models.

\textbf{Model-Agnostic Robustness.} TopoAgent consistently boosts both proprietary and open-weight MLLMs. Notably, it elevates Qwen3-vl's accuracy from 59.3\% to 68.3\%, surpassing the strongest baseline (OctoTools, 64.2\%). This demonstrates that our framework optimizes the reasoning pathway independently of a model's inherent parametric capacity.

\textbf{Breakthroughs in Complex Domains.} TopoAgent excels in rigorous multi-step deduction. On OlympiadBench (Physics), the global average surges from 30.9\% to 41.2\%. Using GPT-5, TopoAgent reaches 60.3\%, far exceeding the 45.8\% baseline and GPT-Functions (40.2\%). These gains highlight how \textit{Adaptive Atomic Fission} prevents execution deadlocks in Olympiad-level problems by dynamically refining task granularity.

\begin{figure}[!t]
    \centering
    \includegraphics[width=0.75\linewidth]{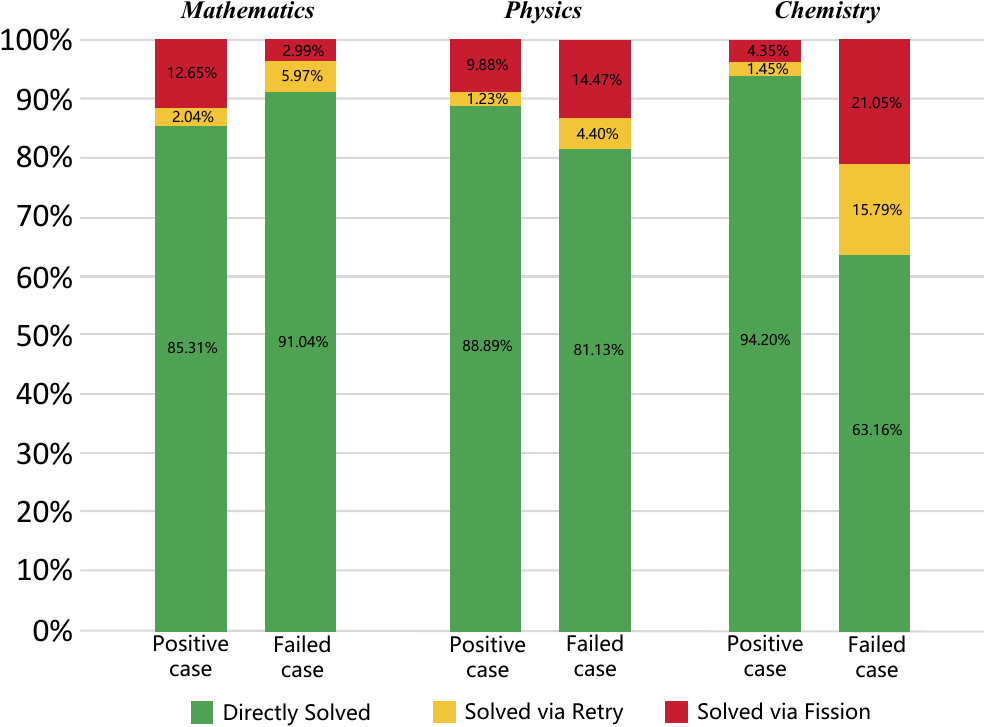}
    \caption{Distribution of atomic task resolution methods by domain and global outcome (Positive vs. Failed). Proportions show atoms resolved directly, via retry, or through Adaptive Atomic Fission.}
    \label{fig:duidiezhu}
\end{figure}

\subsection{Ablation Study}
We ablate TopoAgent's components using the global average performance across all six MLLMs, introducing two variants:

\begin{itemize}
    \item \textbf{w/o DAG Planning:} Replaces DAG orchestration with sequential linear planning, accumulating context step-by-step without topological isolation.
    \item \textbf{w/o Atomic Fission:} Disables dynamic sub-task fracturing, causing execution termination upon tool failure instead of self-evolution.
\end{itemize}

\textbf{Effect of DAG Planning \& Context Isolation.} As shown in Table \ref{tab:ablation}, removing DAG causes a 1.4\% overall drop, with domain-sensitive degradation. Mathematics suffers the most (up to -2.1\%) because it requires extracting independent visual variables (parallel sub-processes); linearizing these accumulates historical noise and dilutes attention. Conversely, Chemistry sees a marginal drop (-0.5\%), as its reasoning inherently follows a sequential, pipeline-like path, making it more tolerant to linear context accumulation.

\textbf{Effect of Adaptive Atomic Fission.} Disabling fission yields a 0.6\% global decrease, with pronounced impacts on dense math and physics (e.g., MathVista -0.9\%, Olympiad -0.6\%). In these domains, complex geometric or kinematic queries frequently exceed external tools' zero-shot boundaries, leading to execution deadlocks. Fission acts as a crucial self-correcting cognitive loop, rescuing bottleneck nodes by dynamically degrading task granularity to ensure robust long-horizon reasoning.

\subsection{Efficacy of Adaptive Atomic Fission}
We analyze the contribution of Adaptive Atomic Fission at the granular task level across three scientific domains. Figure~\ref{fig:duidiezhu} illustrates the resolution distribution of atomic tasks. In successful global outcomes, 12.65\% (Mathematics) and 9.88\% (Physics) of atomic tasks are resolved exclusively via fission after initial failures. This proves that dynamic granularity degradation actively rescues bottleneck nodes rather than acting as a mere theoretical fallback. Conversely, Chemistry relies less on fission, reflecting its inherently sequential, pipeline-like deduction paths.

\begin{figure}[!t]
    \centering
    \includegraphics[width=0.85\linewidth]{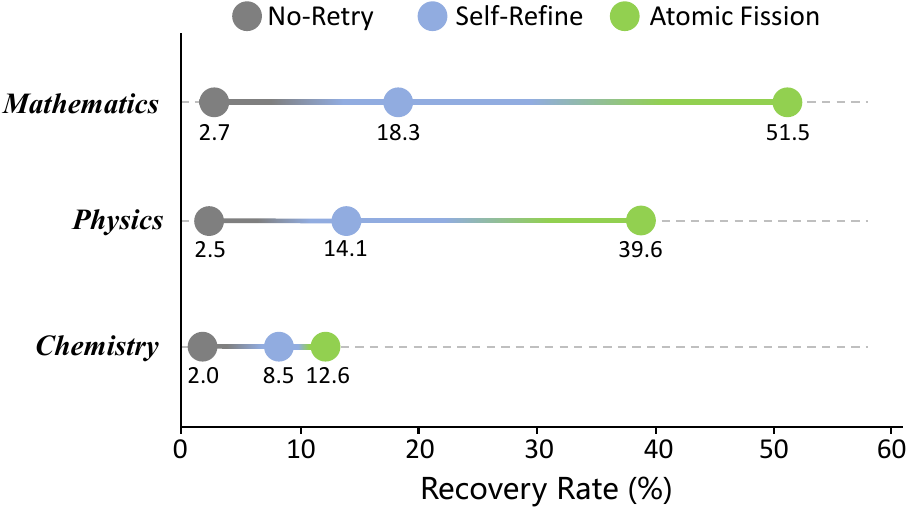}
    \caption{Comparison of recovery rates among different error-handling strategies. Atomic Fission significantly outperforms both No-Retry and traditional Self-Refine strategies across all scientific domains, demonstrating superior self-correction capabilities.}
    \label{fig:yalingtu}
\end{figure}

We further evaluate system resilience against capability boundaries. We compare Atomic Fission with two baselines: (1) \textbf{No-Retry}, which proceeds with missing context, and (2) \textbf{Self-Refine}, which prompts the LLM to blindly retry at the same task granularity.

As shown in Figure~\ref{fig:yalingtu}, standard strategies fail at hard capability boundaries. In Mathematics, Self-Refine recovers only 18.3\% of failed nodes, whereas \textbf{Atomic Fission} achieves a remarkable 51.5\% recovery rate. Instead of blindly retrying monolithic tasks, TopoAgent dynamically fractures bottlenecks to match underlying tool capabilities, ensuring robust long-horizon reasoning.

\begin{figure}[!htbp]
    \centering
    \includegraphics[width=0.85\linewidth]{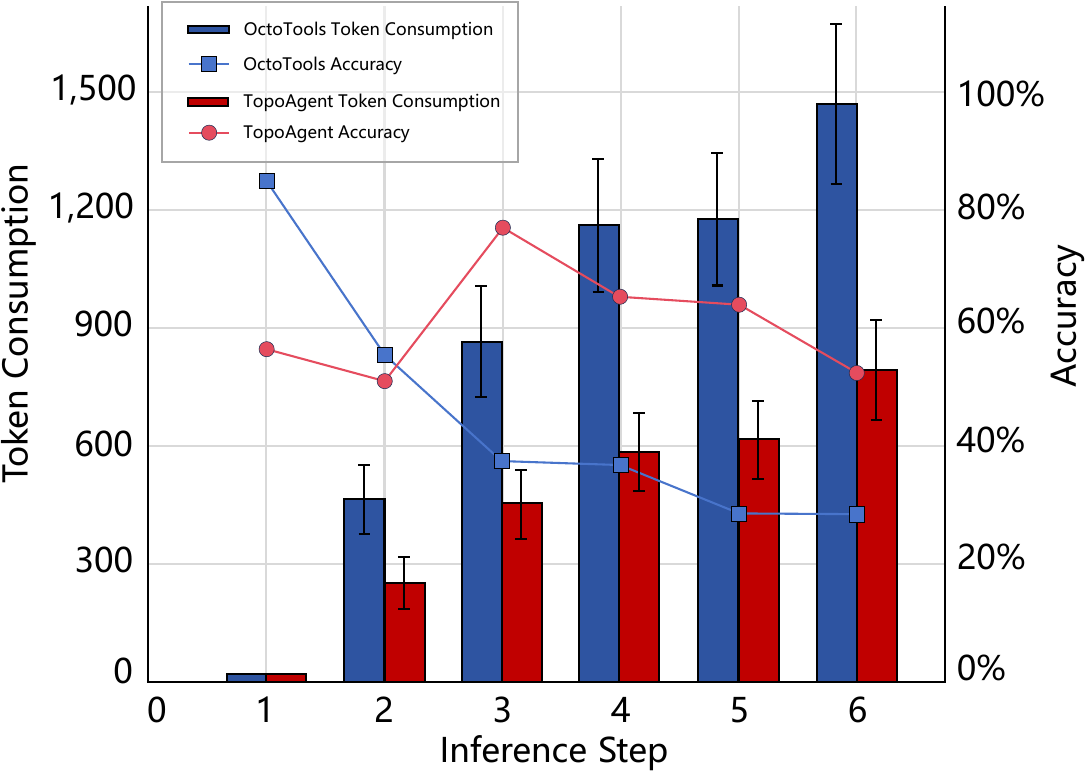}
    \caption{Comparison of token consumption (bar chart) and accuracy (line chart) across inference steps. TopoAgent maintains stable token usage and high accuracy by isolating irrelevant context, whereas the linear baseline suffers from context explosion and performance degradation.}
    \label{fig:token}
\end{figure}

\begin{figure*}[!t]
    \centering
    \includegraphics[width=0.90\linewidth]{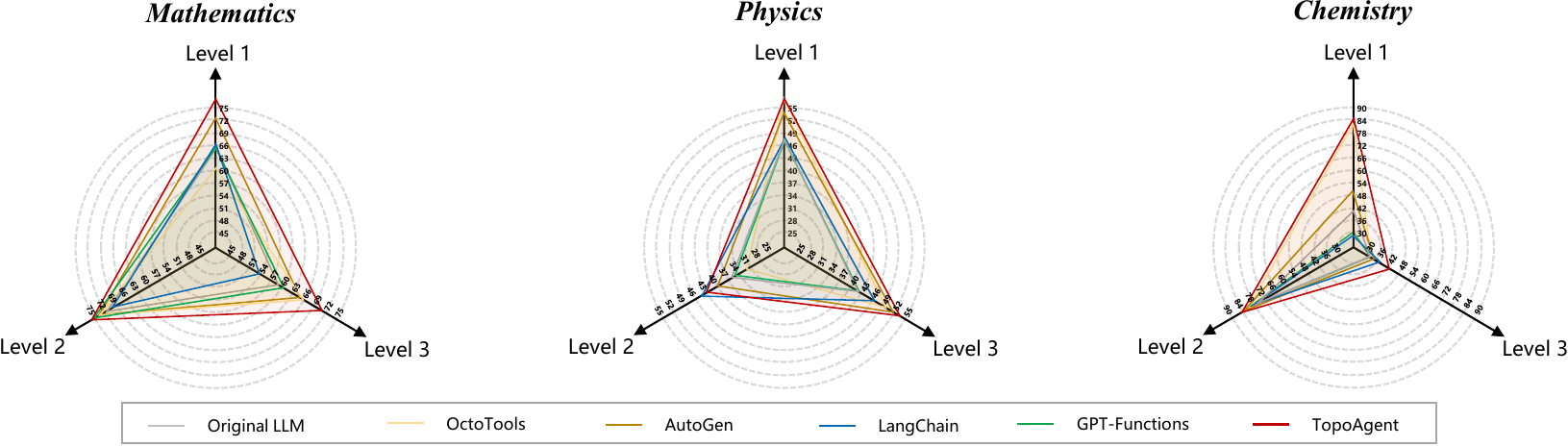}
    \caption{Performance radar charts across cognitive complexities in Mathematics, Physics, and Chemistry. TopoAgent exhibits minimal degradation on highly complex tasks (Level 3), demonstrating superior robustness over other frameworks.}
    \label{fig:leidatu}
\end{figure*}

\begin{figure*}[!t]
    \centering
    \includegraphics[width=0.90\linewidth]{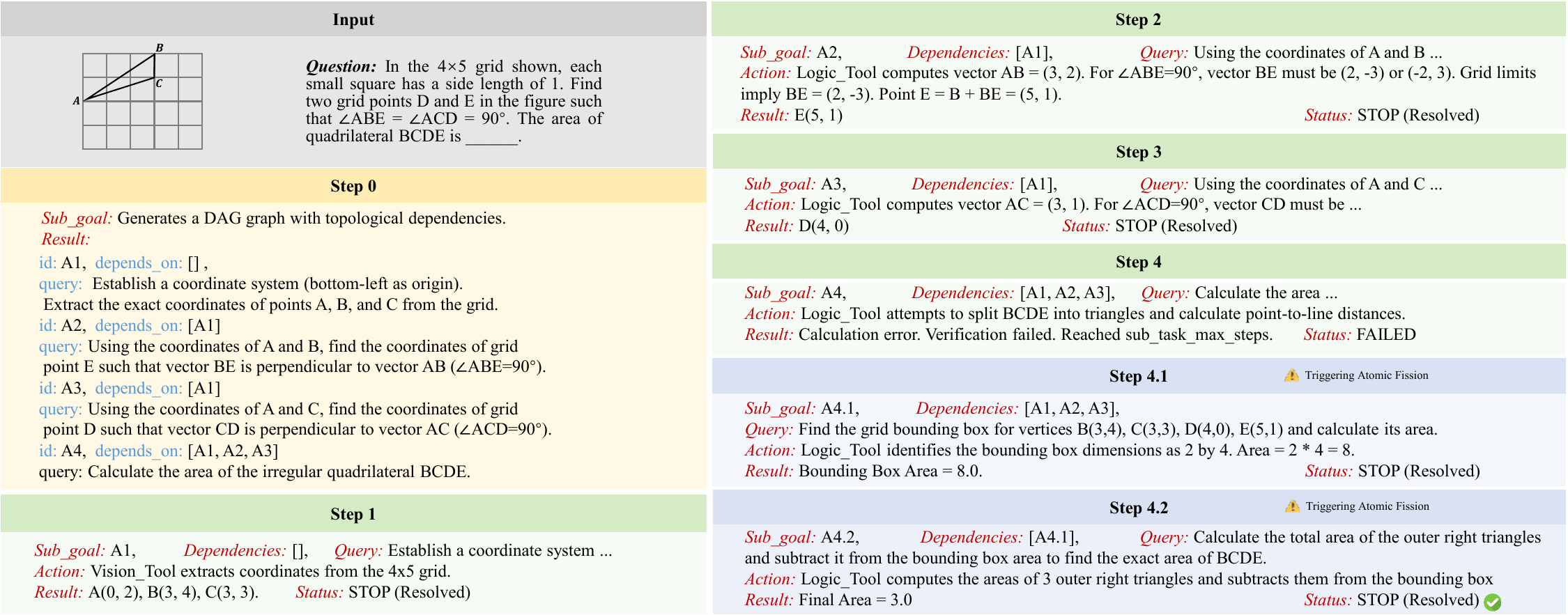}
    \caption{A qualitative case study illustrating the execution trajectory of TopoAgent. The framework first constructs a Directed Acyclic Graph (DAG) for context-isolated atomic execution (Steps 0-3). When a complex reasoning node exceeds the tool's capability boundary (Step 4), the \textit{Adaptive Atomic Fission} mechanism is dynamically triggered, fracturing the bottleneck into finer-grained sub-tasks (Steps 4.1 and 4.2) to ensure robust global resolution.}
    \label{fig:case}
\end{figure*}

\subsection{Context Isolation and Token Efficiency}
To validate the efficiency of DAG planning, we track the average prompt token length and execution accuracy per inference step, comparing TopoAgent against a strong baseline, OctoTools.

As illustrated in Figure~\ref{fig:token}, traditional linear frameworks like OctoTools suffer from a severe "context explosion" as inference deepens. Blindly accumulating historical trajectories causes exponential token growth, leading to attention dilution and a sharp accuracy drop after step 4. Conversely, TopoAgent exhibits remarkable stability. By enforcing strict Context Isolation via DAG topology, TopoAgent injects only highly purified prerequisite states into the current prompt. This keeps token consumption low regardless of reasoning depth and shields the reasoning engine from irrelevant historical noise. Consequently, TopoAgent effectively mitigates long-context hallucinations and sustains high accuracy even at deep inference steps, proving that topological routing is both computationally efficient and crucial for reliable long-horizon reasoning.

\subsection{Robustness Across Cognitive Complexities}
To evaluate robustness against problem complexity, we adopt the cognitive evaluation taxonomy introduced by ~\cite{li2026cognitive}, which categorizes tasks into three progressive levels: Level 1 (perception), Level 2 (reasoning), and Level 3 (critical thinking). Figure \ref{fig:leidatu} reveals that while baseline agents achieve comparable performance on basic visual extraction (Level 1), they suffer severe degradation on Levels 2 and 3 due to error accumulation and context dilution in deep reasoning chains. Conversely, TopoAgent exhibits minimal performance drop on Level 3 tasks across all domains. This sustained robustness directly validates our core designs: DAG-based context isolation preserves logical consistency over long horizons, while Adaptive Atomic Fission dynamically degrades task granularity to prevent the execution deadlocks that typically paralyze standard sequential planners.

\subsection{Qualitative Analysis of Dynamic Evolution}
\label{sec:case_study}

To qualitatively elucidate the internal mechanics and self-correcting capabilities of TopoAgent, Figure \ref{fig:case} presents an execution trajectory on a complex reasoning task.

Rather than attempting a monolithic, end-to-end generation, the framework initially employs the Decomposer to fracture the overarching query into a Directed Acyclic Graph (DAG) of visually-grounded atomic tasks (Step 0). This topological formulation explicitly defines prerequisite dependencies. By doing so, TopoAgent strictly enforces context isolation during Steps 1 through 3; each reasoning node is fed only the deterministic outputs of its direct ancestors, effectively shielding the reasoning engine from irrelevant historical noise and visual-semantic misalignment.

Crucially, the trajectory highlights the system's resilience when confronting inherent capability boundaries. During the execution of a complex terminal node (Step 4), the assigned tool encounters a cognitive bottleneck and fails to resolve the task within the prescribed step limit. In standard linear agentic frameworks, such an impasse would inevitably precipitate a global execution deadlock. In stark contrast, TopoAgent autonomously triggers the \textit{Adaptive Atomic Fission} mechanism. It dynamically analyzes the failure trace and fractures the monolithic bottleneck node into finer-grained, executable sub-atoms (Steps 4.1 and 4.2).

By degrading the task granularity on the fly, the agent successfully circumvents the execution bottleneck and derives the correct final state without requiring a global recompilation of the reasoning graph. This paradigm of ``plan, execute, and dynamically evolve'' empirically demonstrates how TopoAgent effectively bridges the gap between static task planning and the dynamic capability limits of underlying MLLMs, establishing a highly robust pathway for long-horizon scientific discovery.

\section{Conclusion}

We present TopoAgent, a self-evolving topological framework for multimodal scientific reasoning that addresses visual-semantic misalignment, long-context hallucinations, and static execution failures in conventional linear agentic systems. Through Visually-Grounded Atomic Decomposition, DAG-based planning, and Adaptive Atomic Fission, TopoAgent enables context-isolated reasoning and dynamic execution adaptation. Extensive experiments on scientific benchmarks in Mathematics, Physics, and Chemistry demonstrate state-of-the-art performance, strong token efficiency, model-agnostic robustness, and reliable reasoning on complex scientific tasks. These results highlight TopoAgent as a scalable framework for autonomous scientific reasoning.

\bibliography{aaai2027}


\end{document}